\documentclass{elsarticle}

\usepackage{lineno,hyperref}

\usepackage[utf8]{inputenc}
\usepackage{amsmath}
\usepackage{mathtools}
\usepackage{algorithm}
\usepackage{algpseudocode}
\usepackage{blindtext}
\usepackage{tabularx}
\usepackage[english]{babel}
\usepackage{array}
\usepackage{multirow}
\usepackage{booktabs}
\usepackage{lscape}
\usepackage{xr-hyper}
\usepackage{tablefootnote}
\usepackage{graphicx}
\usepackage{siunitx}
\usepackage{comment}
\usepackage{fancyvrb}
\usepackage[dvipsnames]{xcolor}

\usepackage{todonotes}

\journal{Journal of \LaTeX\ Templates}

\begin{document}

\begin{frontmatter}
\title{A Neuro-Symbolic Explainer for Rare Events:\\
A Case Study on Predictive Maintenance}

\author[fep,inesctec]{Jo\~ao Gama\corref{mycorrespondingauthor}} 
\author[fcup,inesctec]{Rita P. Ribeiro}
\author[icmc,inesctec]{Saulo Mastelini}
\author[inesctec]{Narjes Davari}
\author[fep,inesctec]{\\Bruno Veloso} 

\address[fep]{Faculty of Economics, University of Porto, 4200-464 Porto. Portugal}
\address[fcup]{Faculty of Sciences, University of Porto, 4169-007 Porto, Portugal.}
\address[icmc]{ICMC - University of S. Paulo, 13566-590 São Carlos, Brazil}
\address[inesctec]{INESC TEC, 4200-465
Porto, Portugal.}



\begin{abstract}
Predictive Maintenance applications are increasingly complex, with interactions between many components. 
Black-box models are popular approaches based on deep-learning techniques due to their predictive accuracy. 
This paper proposes a neural-symbolic architecture that uses an online rule-learning algorithm to explain when the black-box model predicts failures.
The proposed system solves two problems in parallel: 
{\it i)} anomaly detection and {\it ii)} explanation of the anomaly.
For the first problem, we use an unsupervised state-of-the-art autoencoder. 
For the second problem, we train a rule learning system that learns a mapping from the input features to the autoencoder's reconstruction error. Both systems run online and in parallel.
The autoencoder signals an alarm for the examples with a reconstruction error that exceeds a threshold. 
The causes of the signal alarm are hard for humans to understand because they result from a non-linear combination of sensor data. The rule that triggers that example describes the relationship between the input features and the autoencoder's reconstruction error. 
The rule explains the failure signal by indicating which sensors contribute to the alarm and allowing the identification of the component involved in the failure. 
The system can present global explanations for the black box model and local explanations for why the black box model predicts a failure. 
 We evaluate the proposed system in a real-world case study of Metro do Porto and provide explanations that illustrate its benefits.

\end{abstract}

\begin{keyword}
Explainable AI \sep 
Online Anomaly Detection
\sep Predictive Maintenance
\MSC[2010] 00-01\sep  99-00
\end{keyword}

\end{frontmatter}


\section{Introduction}
Real-world predictive maintenance applications are increasingly complex, with extensive interactions of many components. 
Data-driven predictive maintenance (PdM) solutions are a trendy technique in this domain, and especially the black-box models based on deep learning approaches show promising results in fault detection accuracy and capability of modelling complex systems~\cite{Khoshafian2020,Serradilla2022}.
However, the decisions these black-box models make are often difficult for human experts to understand and make the correct decisions. 
The complete repair plan and maintenance actions that must be performed based on the detected symptoms of damage and wear often require complex reasoning and planning processes involving many actors and balancing different priorities. It is unrealistic to expect this complete solution to be created automatically – too much context needs to be considered. Therefore, operators, technicians, and managers require insights to understand what is happening, why, and how to react. 
Today's primarily black-box machine learning models do not provide these insights, nor do they support experts in making maintenance decisions based on detection deviations. 
The effectiveness of the PdM system depends much more on the pertinence of the actions operators perform based on the triggered alarms than on the accuracy of the warnings themselves.

Fault detection is one of the most critical components of predictive maintenance. Nevertheless, predictive maintenance goes far behind predicting a failure, and it is essential to understand the consequences and the collateral damages of the failure.
A critical issue in predictive maintenance applications is designing the maintenance plan after a fault is detected or predicted. To elaborate the recovery plan, it is essential to know the causes of the problem (root cause analysis), which component is affected, and the expected remaining useful life of the equipment. 
Explanations in predictive maintenance play a relevant role in identifying the causes of failure, e.g., the component in failure. 
This is the type of information required to define the repair plan.

The main contribution of this work is an online explanatory layer for deep-learning-based anomaly detection systems in data-driven predictive maintenance. We propose a two-layer architecture. The first layer uses an unsupervised deep-learning model for detecting failures. The second layer learns interpretable models (regression rules in the current implementation) that explain the outputs of the detection layer model.
Several works discuss the problem of explainability of deep learning models, for example, \cite{molnar2022,GuidottiMRTGP19,hall2019,Moreira2021AnIP}. 
To our knowledge, none of the works focuses on explaining deep-learning models for the rare cases: failures. That is the focus of this work.
In the experimental section, we analyze 
a real-world case study from public transport services, showing the relevance of the explanations for predictive maintenance.

This paper is organized as follows. The following section briefly introduces the data-driven predictive maintenance (PdM) problem and the importance of obtaining explanations for detected faults. We also refer to existing works in the field of Explainable AI.
Section~\ref{sec:approach} introduces our two-layer architecture for explainable PdM. 
Then, in Section~\ref{sec:casestudy1}, we present the case study: the Metro do Porto problem.
We conclude in Section~\ref{sec:concl}, pointing out the generality of our approach to other similar scenarios and further research directions.

\section{Background and Related Work}
\label{sec:background}
Maintenance is the process that deals with the health of equipment and system components to ensure their normal functioning under any circumstances. Over the years, and due to technological advances, different maintenance strategies have been developed. Deep-learning techniques are quite popular, primarily due to their performance in detecting anomalies and failures ~\cite{JapkowiczMG95,Serradilla2022}.

In a well-known PdM story, a train entered the workshop due to an alarm in the compressed air unit. They replaced the wrong part in the workshop, and the train was returned to operation. A few hours later, it was forced to return to the workshop with the same problem. This persistent problem in maintenance reinforces the need to explain fault alarms in an understandable way for everyone.


The state-of-the-art machine learning literature identifies several existing interpretable models~\cite{GuidottiMRTGP19}: decision trees, regression rules, linear models, some Bayesian Networks, etc. 
These models are considered easily understandable and interpretable for humans.
In \cite{GuidottiMRTGP19} the authors differentiate between {\it global} and {\it local} explainability: 
\begin{quote}
"A model may be completely interpretable, i.e., we can
understand the whole logic of a model and follow the entire reasoning leading to all the different
possible outcomes. In this case, we are speaking about global interpretability. Instead, we indicate
with local interpretability the situation in which it is possible to understand only the reasons for
a specific decision: only the single prediction/decision is interpretable."
\end{quote}
In our case, we use online regression rules to generate global and local models of interpretability.

\section{From Fault Detection to Anomaly Explanation}\label{sec:approach}

One approach in data-driven predictive maintenance (PdM) is to learn a system's normal operating condition so that any significant deviation from that working condition can be spotted as a potential failure. One possibility is to attain this by learning LSTM-AE on normal operation examples. The intuition is that if we give an abnormal example of operation, the reconstruction error of the LSTM-AE will be very high. Those examples are relevant from the PdM perspective and are typically rare.

The purpose of online anomaly detection and explanation for fault prediction is to identify and describe the occurrence of defects in the operational units of the system that result in unwanted or unacceptable behaviour of the whole system. As reported, a single on-the-road breakdown substantially outweighs the cost of a single unneeded component replacement. This work implements an anomaly detection framework based on deep learning as a reliable and robust technique that issues an alert whenever a sequence of abnormal data is detected. However, having the best experimental accuracy in the anomaly detection segment is no longer enough. Explaining the obtained outputs to users is necessary to increase their confidence. To this end, in parallel with anomaly detection, the anomalies are described based on rule learning algorithms.

\begin{figure}[!bt]
\includegraphics[width=1\linewidth]{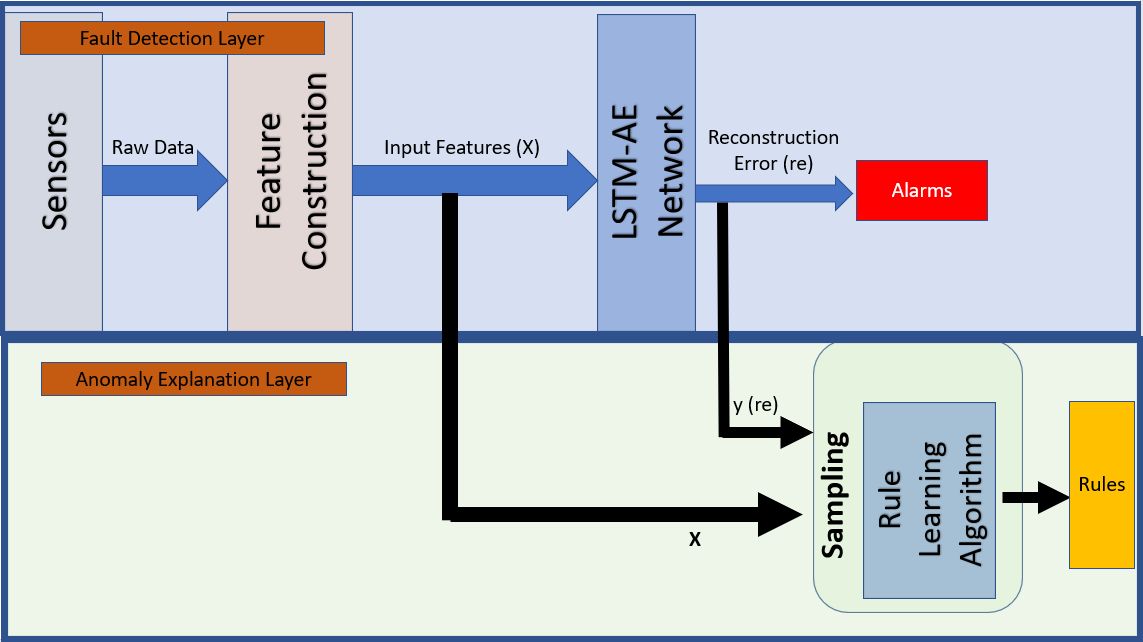}
\caption{Global Architecture of the Online Anomaly Explanation System. The top panel details the fault detection system, while the bottom details the explanation system. Both systems run online and in parallel.}
\label{fig:fig0}
\end{figure}

Figure~\ref{fig:fig0} presents the global architecture of the proposed online anomaly explanation system.
It is a neuro-symbolic system~\citep{Sarker2021} where a rule learning system learns a surrogate model of the LSTM-AE.
It works online, receiving real-time data from sensors installed in the machine (train, bus, complex equipment, etc.) we want to monitor.
It signals alarms when it receives data that does not correspond to the machine's normal behaviour.  
The system also explains which features contributed more to the signal alarm.

The figure details the two layers of the system: the {\em fault detection} and {\em explanation} layers.
The main component in the {\em detection layer} is the LSTM-AE network trained with examples from the normal behaviour of the machine we want to monitor. Each new observation is propagated through the autoencoder network. 
The reconstruction error is computed as the mean square difference between the input and the output:  $re = 1/n \times \sum_{i=1}^n(x_i-\hat{y_i})^2$, where $n$ is the number of features, $x_i$ the input features, and $\hat{y_i}$ the corresponding output. When the reconstruction error, $re$, exceeds a threshold, we signal an alarm, meaning that the input is not from the system's normal behaviour.  
The second layer, the {\em explanation layer}, receives as independent variables ({\bf $X$}) the input features of the LSTM-AE network, and the dependent variable ($y = re$) is the corresponding reconstruction error. It learns an interpretable model, a mapping $y=f(X)$. Both layers run online and in parallel, which means that for each observation $X$, the model propagates the observation through the network and recomputes the reconstruction error ($y=re$). The second layer updates its model using $\langle X,y \rangle$. If $y > threshold$, the system signals an alarm with an explanation of the rules that trigger, for example, $X$.
The first layer is the failure detection layer and is unsupervised. The second layer is supervised and explains the failure regarding the input features.

We assume that high extreme values of the reconstruction error ($re$) are a potential indicator of failures.
This architecture allows two explanations: i) Global level, the set of rules learned that explains the conditions to observe high predicted values, and ii) Local level, which rules triggered for a particular input.

An essential characteristic of this problem is that failures are rare events. An LSTM-AE trained with normal operating conditions is expected to struggle to reconstruct the input when it receives an abnormal operating condition. Thus, the reconstructed error is expected to be higher for such cases.

We aim to obtain explanations focused on cases 
with high and extreme reconstruction errors, which can indicate a potential failure. 
This particularity of having a non-uniform preference across the target variable domain and focusing on scarcely represented values configures an imbalanced learning problem~\cite{RibeiroMoniz20}.
As reported in many related studies in the area~\cite{BrancoTR16}, standard machine learning approaches and evaluation metrics hinder performance in rare cases, which might be of utmost importance. Thus, they are not suitable for addressing predictive tasks from imbalanced domains. From the learning perspective, two main approaches for tackling this type of task exist at the data level and algorithm level.
This work resorted to data-level approaches applied to the rule learning algorithm and used a specific error metric to cope with imbalanced domain learning.
In the following subsections, we first describe the online anomaly detection and then explain the online rule learning algorithm, the applied sampling techniques, and the used evaluation metric to properly assess the effectiveness of our approach in an imbalanced data streams regression scenario~\cite{AminianRG21}.
\subsection{Fault detection based on LSTM-AE}\label{sec:AnomalyDetect}
\begin{figure}[!htb]
\includegraphics[width=1\linewidth]{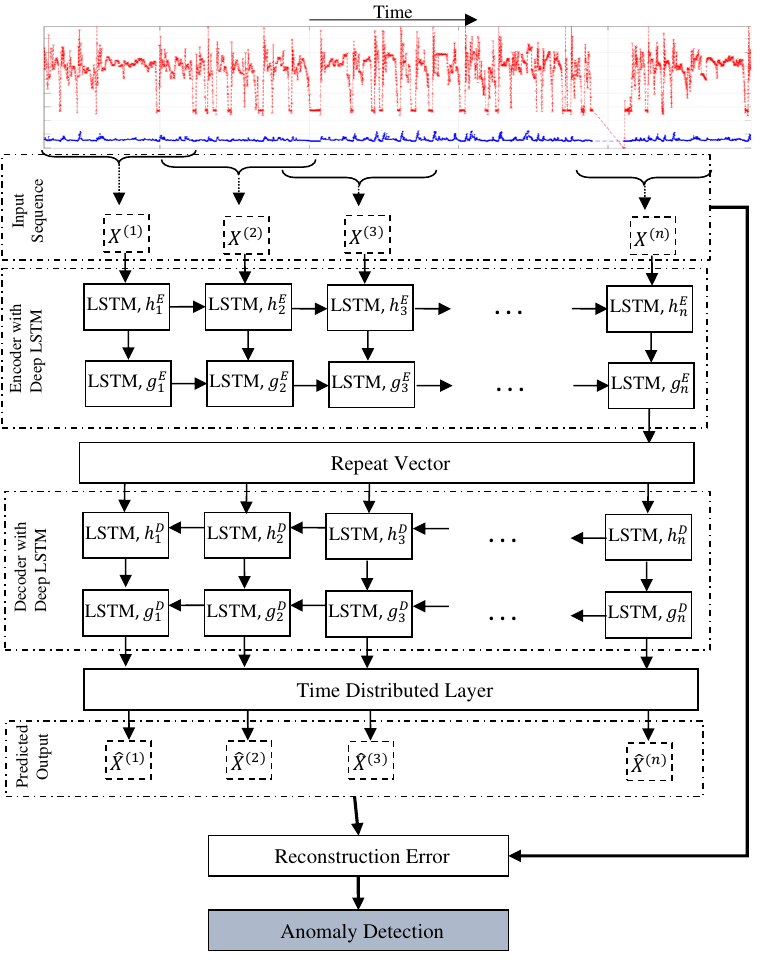}
\caption{Schema of deep LSTM-AE for anomaly detection}
\label{fig:LSTMAE}
\end{figure}

We use the system's normal behaviour data in the fault detection framework to train an initial LSTM-AE network offline to capture the time series' temporal characteristics. 
In the initialization phase, we define the threshold for signal alarms. We apply a boxplot, a standard method to detect univariate outliers, to the reconstruction error. The threshold is defined as $Q_3 + 3 \times IQR$. 
After the initialization phase,  the trained network is used online.

The APU activity is controlled by the compressor (COMP). It has two distinct states: charging the pneumatic circuit (COMP has value 0) or consuming air (COMP has value 1). A compressor cycle is a full sequence of charging the pneumatic circuit and consuming air, starting from the timestamp when the COMP signal changes to 0 and ending at the timestamp preceding the following change of the COMP signal to 0. The compressor cycle defines the window for the LSTM-AE.

The overview of the deep LSTM-AE model, which consists of two modules, LSTM encoders and LSTM decoders, is illustrated in Figure~\ref{fig:LSTMAE}. LSTM encoders are generally used to receive the input sequences, extract and compress features from inputs through nonlinear transformation, and then pass the high-level features of the data as the latent vectors on the hidden state of the last time step of the LSTM decoder. The repeat vector layer is applied to repeat the context vector received from the encoder and feed it to the decoder as an input. The LSTM decoder creates the reconstructed data at the sliding time window \textit{i} by applying the hidden decoding state at the sliding time window \textit{i} and the predicted time series at the sliding time window $(i-1)$. 
A time-distributed dense layer is employed in the output of the LSTM decoder to maintain one-to-one relationships between input and output and generate specific output for each time window~\cite{nguyen2021forecasting}.

The reconstruction error, computed as the divergence between the input features and the reconstructed one, is applied to update the model parameters and compute the anomaly score for each sliding window. The root mean square of the reconstruction error $\text{RMS}_{re}$ for the training dataset is used to estimate a threshold value ($thr_{re}$) through a boxplot analysis to label data. The boxplot is a consistent method to display the distribution of the dataset, which can be used to ignore extreme observations. The boxplot analysis of RMSE of training data was previously proposed in~\cite{ribeiro2016sequential} for obtaining the threshold value. 
If the $\text{RMS}_{re}$ of a time window $i$ is larger than the threshold $(i-1)$, then the window $i$ is considered an anomalous observation. 
Otherwise, it is regarded as normal behaviour. 
Normal data is used to retrain the LSTM-AE model and to update the threshold $thr_{re}$. 
Adaptive threshold estimation can follow the new data distribution in that time window. 
Finally, the output of the labelling step is inputted into a post-processing phase to decrease the false alarm rates~\cite{ribeiro2016sequential}. 
To this end, a low-pass filter removes the network output's sudden variations and unwanted components. 
A single-time window alarm is insufficient to demonstrate a persistent failure. 
Only a sequence of abnormal sliding windows is an indication of failure.

\subsection{The Adaptive Model Rules Algorithm}
Decision rules are one of the most expressive and interpretable models for machine learning~\cite{GuidottiMRTGP19}.
The Adaptive Model Rules (AMRules)~\cite{DuarteGB16} was the first stream rule learning algorithm for regression problems
~\footnote{Nowadays, there are other streaming regression algorithms; a good survey is~\cite{Krawczyk2022}. One of the reasons why we use AMRules is that the code is publicly available in RIVER~\cite{2020river}}.
In AMRules, the antecedent of a rule is a conjunction of conditions on the attribute values. The consequent can be a constant (the mean of the target variable) or a linear combination of the attributes. 

The set of rules learned by AMRules can be ordered or unordered. 
They employ different prediction strategies to achieve {\it 'optimal'}  prediction. 
In the case of ordered rules, only the first rule that covers an example is used to predict the target value. 
In the case of unordered rules, all rules covering the example are used for prediction, and the final prediction is decided by aggregating predictions using the mean.

Each rule in AMRules implements three prediction strategies: 
{\it i)} the mean of the target attribute computed from the examples covered by the rule;
{\it ii)} a linear combination of the independent attributes;
{\it iii)} an adaptive strategy that chooses between the first two strategies, the one with the lower mean absolute error (MAE) in the previous examples. In this case, the MAE is computed following a fading factor strategy.
We use the implementation of AMRules available in ~\cite{2020river}.

The standard AMRules algorithm learns rules to minimize the mean squared error loss function. In the predictive maintenance context, the relevant cases are failures, rare events characterized by large values of the reconstruction error. 
We aim to apply AMRules to predict the reconstruction error obtained from the original set of features as the target value.
The most critical cases that constitute potential failures have high and extreme target values. 
Even though they are rare, we want our model to focus on them.
The following section explains two sampling strategies to learn from rare events.

\subsection{Online Sampling Strategies for Extreme Values Cases}
Few approaches to learning from imbalanced data streams discuss the task of regression. 
In this work, we use the approach described in ~\cite{AminianRG21}, which resorts to Chebyshev's inequality as a heuristic to disclose the type of incoming cases (i.e., frequent or rare).
It guarantees that "nearly all" values are close to the mean in any probability distribution.
More precisely, no more than $\frac{1}{t^2}$ of the distribution's values can be more than $t$ times standard deviations away from the mean. 
This property substantiates the fact that extreme values are rare.
Although conservative, the inequality can be applied to entirely arbitrary distributions (unknown except for mean and standard deviation).

Let  $Y$ be a random variable with finite expected value $\overline{y}$ and finite non-zero standard deviation $\sigma$. 
Then, Chebyshev's inequality states that  for any real number $t > 0$, we have:

\begin{equation}
 \Pr(|y-\overline{y}|\geq t\sigma) \leq \frac{1}{t^2}
 \label{equa1}
\end{equation} 

Only the case $t > 1$ is helpful in the above inequality. 
In cases  $t < 1$, the right-hand side is greater than one, and thus, the statement will be {\it always true} as the probability of any event cannot be greater than one. Another ``always true" case of inequality is when $t = 1$. In this case, the inequality changes to a statement saying that the probability of something is less than or equal to one, which is ``always true".

For $t=\frac{|y - \overline{y}|}{\sigma}$ and $t > 1$, we define \textit{frequency} score of observation $\langle x, y\rangle$ as:

\begin{equation}
P(\mid \overline{y} - y \mid \geq t) = \frac{1}{\left(\frac{|y - \overline{y}|}{\sigma}\right)^2}\label{equa2}
\end{equation}

The above definition states that the probability of observing $y$ far from its mean is small and decreases as we get farther away from the mean.

In an imbalanced data stream regression scenario, considering the mean of target values of the examples in the data stream ($\overline{y}$), examples with rare extreme target values are more likely to occur far from the mean. In contrast, examples with frequent target values are closer to the mean. So, given the mean and standard deviation of a random variable, Chebyshev's inequality can indicate the degree of the rarity of an observation. 
Its low and high values imply that the observation is probably a frequent or a rare case, respectively.

Having been equipped with the heuristic to discover if an example is rare or frequent, the next step is to use such knowledge in training a regression model.
In~\cite{AminianRG21}, authors proposed ChebyUS and ChebyOS, respectively, under-sampling and over-sampling methods. However, the most effective method in the predictive maintenance scenario is over-sampling. We aim to provide the learning algorithm with more abnormal examples, i.e. those with high extreme reconstruction error, as they might describe potential failures.

 \subsubsection{ChebyOS: Chebyshev-based Over-Sampling}
One method to balance a data stream is to over-sample rare cases of the incoming imbalanced data stream. Since those rare cases in data streams can be discovered by Chebyshev's probability, they can be easily over-sampled by replication.

For each example, a $t$ value can be calculated by Equation \ref{eq:eq_compute_t_over_sampling} which yields the result in $[0\;\;+\infty)$.
\begin{equation}
  t= \frac{|y - \overline{y}|}{\sigma}
  \label{eq:eq_compute_t_over_sampling}
\end{equation}

While the $t$ value is small for examples near the mean, it would be larger for ones farther from the mean and has its largest value for examples located in the farthest distance to the mean (i.e., extreme values). In their proposal~\cite{AminianRG21}, authors limit the function in Equation \ref{eq:eq_compute_t_over_sampling} to produce only natural numbers as follows:

\begin{equation}
  K =  \left \lceil \frac{|y - \overline{y}|}{\sigma} \right \rceil
  \label{eq:eq4}
\end{equation}

$K$ is expected to have greater numbers for rare cases. In this over-sampling method, the $K$ value is computed for each incoming example, meaning that the example is presented exactly $K$ times to the learning algorithm.
Examples not as far from the mean as the standard deviation are probably frequent cases. They contribute only once in the learner's training process while the others contribute more times~\footnote{The implementations of {\tt ChebyOS} and {\tt ChebyUS} used are available in River~\cite{2020river}.}.

\subsection{Evaluation}
As stated, errors over rare cases in imbalanced domains are more costly, as they are the most relevant ones from the domain perspective. 
In this setting, the relevant cases are the ones for which the reconstruction error is abnormally high, meaning that the autoencoder has struggled to reproduce the input. These are potential indicators of failure.
With this aim, we use the relevance function $\phi: \mathcal{Y} \rightarrow [0,1]$ introduced in~\cite{Ribeiro2011} as a weight function to each example's target in the Root Mean Square Error (RMSE).
We use the proposed automatic method~\cite{Ribeiro2011,RibeiroMoniz20} based on a boxplot to obtain the relevance function that specifies values in the tail(s) of the distribution of the target variable as regions of interest.

This case study focuses on high-extreme values, representing potential failures. Thus, the interpolation will be made on the three control points: minimum and median values with a minimum relevance of 0 and the upper adjacent value of the boxplot with a maximum relevance of 1.
The interpolation function guarantees a smooth increasing function within the $[0,1]$ domain, being constant outside the range of target values supplied.  
        
As an evaluation metric, we have used both the standard 
RMSE and the $\text{RMSE}_\phi$ - $\phi()$ weighted version of this function where the prediction error for each example is multiplied by the relevance value assigned to that  example, as follows:
    
\begin{equation}
\text{RMSE}_{\phi} = \sqrt{\frac{1}{n}\sum_{i=1}^{n}{\phi(y_i)\times(y_i-\hat{y}_i)^2}},
\label{eq:eq6}
\end{equation}

\noindent where $y_i$, $\hat{y}_i$ and $\phi(y_i)$ refer to the true value, the predicted value and the relevance value for $i$-th example in the data set, respectively. We used sliding windows of length $1000$ to compute the boxplot needed to calculate the relevance function. Therefore, we report RMSE$_\phi$ values computed for each window of instances.

% ...



\begin{figure}[!htb]
\includegraphics[width=1\linewidth]{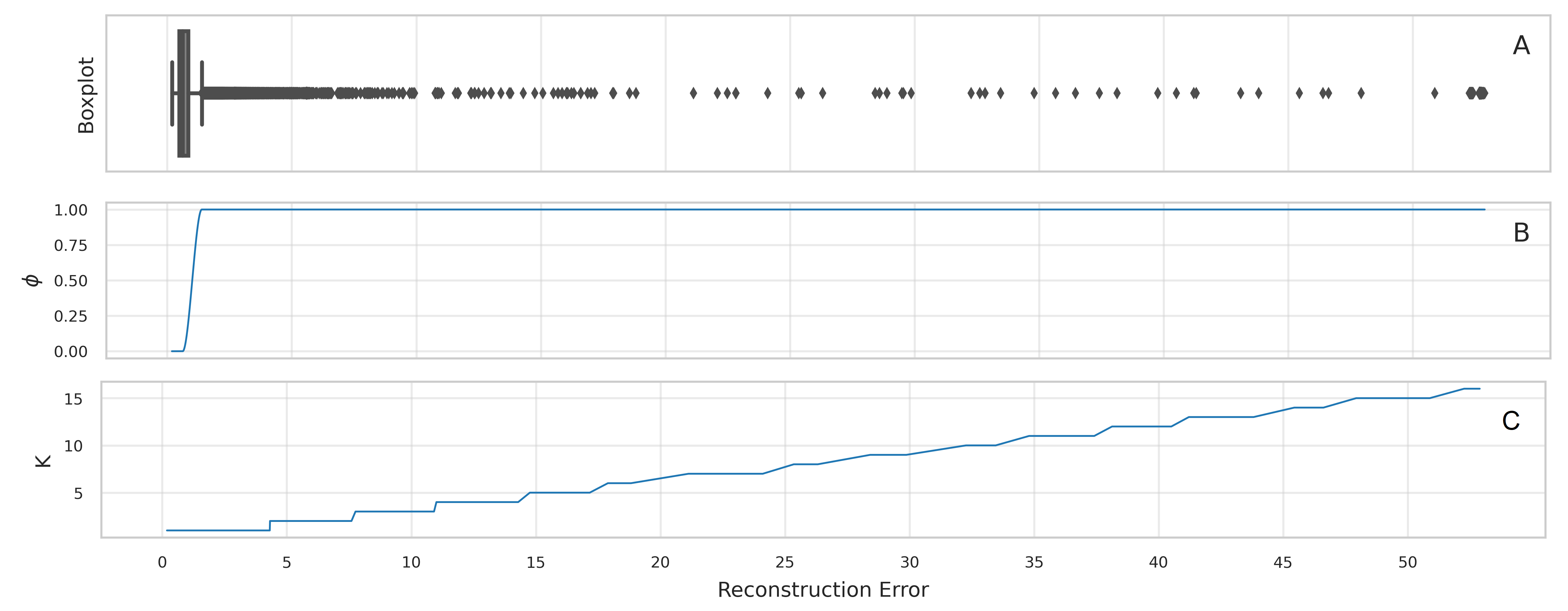}
\caption{A) Box plot for the target variable - reconstruction error; B) Relevance $\phi(.)$ of the target values; 
C) K-value used in the over-sampling approach.}
\label{fig:fig2}
\end{figure}

Figure~\ref{fig:fig2} depicts the summary of values associated with a numeric variable according to its distribution. In particular, 
we have 
the relevance function $\phi()$ derived automatically from the boxplot and
the computed $K$  value used in the over-sampling approach (ChebyOs).

\section{The Metro do Porto Case Study }
\label{sec:casestudy1}

\begin{figure}[!hbt]
    \centering
    \includegraphics[width=0.95\textwidth]{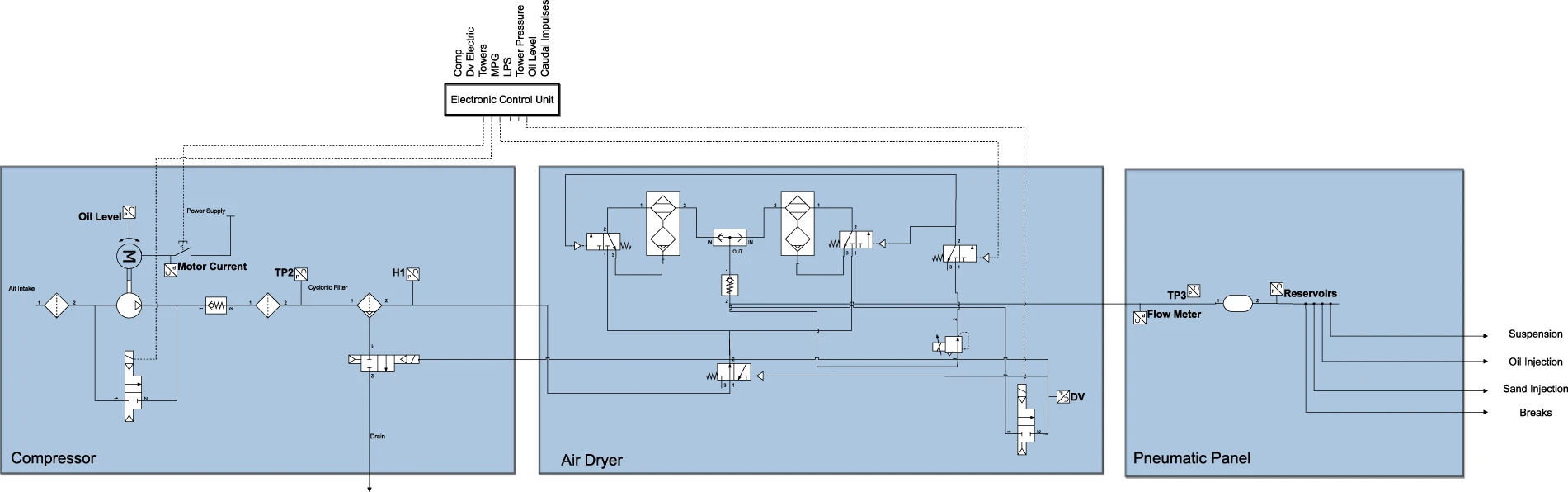}
    \caption{The Air Production Unit system with the position of
the main sensors \cite{veloso2022metropt}.}
    \label{fig:apu}
\end{figure}

This case study includes a single dataset \cite{veloso_bruno_2022_6854240} collected from a  single APU installed on a train, see Figure~\ref{fig:apu}. 
The dataset comprehends signals from eight analogue and eight digital sensors for six months, from January to June 2022, as described in Table \ref{tab:msensors}. 
The detailed description of the data collection process appears in \cite{veloso2022metropt}.

\begin{table}[!htb]
    \centering
    \small
    \caption{Onboard analog and digital sensors from APU train \cite{barros2020failure,veloso2022metropt}.}
    \label{tab:msensors}
    \begin{tabular}{rrl}
        \toprule
        Nr. & Module & Description\\
        \midrule
        \multicolumn{3}{l}{Analog}\\
        \midrule
        1 & Compressor & TP2 - Compressor Pressure\\
        2 & Air Control Panel & TP3 - Pneumatic panel Pressure\\
        3 & Air Control Panel & H1 - Pressure above 10.2 Bar \\
        4 & Air Dryer & DV - Air Dryer Tower Pressure \\
        5 & Air Control Panel & Reservoirs - Pressure  \\
        6 & Compressor & Oil Temperature \\
        7 & Air Control Panel & Flow meter \\
        8 & Compressor & Motor Current\\
         \midrule
        \multicolumn{3}{l}{Digital}\\
        \midrule 9 & Electronic Control Unit & COMP - Compressor on/off \\
        10 & Electronic Control Unit & DV electric - Compressor outlet valve\\
        11 & Electronic Control Unit &  Towers - Active tower number \\
        12 & Electronic Control Unit & MPG - Pressure below 8.2 Bar\\
        13 & Electronic Control Unit &  LPS - Pressure is lower than 7 bars \\
        14 & Electronic Control Unit & Towers Pressure\\
        15 & Compressor & Oil Level - Level below min \\
        16 & Air Control Panel & Caudal impulses\\
        \bottomrule
    \end{tabular}
\end{table}

Each sensor data was recorded with a 1 Hz sampling rate while the compressor was running on the train. The dataset contains two types of failures, one related to air leaks and another regarding oil leaks, and we received the maintenance reports confirming these failures \cite{veloso2022metropt}. 
We now report the features extracted from the onboard sensor data and the outcomes of our proposed methodology.

\subsection{Feature Construction}

In this use case, we extract the following features: 
\begin{enumerate}
\item the number of ones in one cycle of working time compressor (T\_run+T\_idle) for all digital sensors; 
\item the bins for TP2, TP3, H1, Flowmeter, Motor Current; 
\item the {min} and {max} of oil temperature, {DV} pressure, and reservoirs;  
\item the run and idle time of the compressor.
\end{enumerate}

The bin construction is separated into two moments i) the fill of the tanks where we construct two bins, B1 and B2, and ii) the bins B3 until B7, which occurs during the emptying of the tanks.
All the feature construction and failure detection processes used in this case study are described in detail in \cite{davari2021predictive}.

\subsection{Experimental setup}

The components used to create the processing pipelines are available in \href{https://riverml.xyz}{River}. 
In all the cases, we selected the same hyperparameters for AMRules to provide a fair comparison.
The interval between split attempts, $n_{min}$, was set to $100$.
The split confidence parameter, $\delta$, was set to $0.05$, and the selected prediction strategy was the target mean.
We relied on the Truncated Extended Binary Search Tree (TE-BST)\cite{mastelini2021using} as a feature splitter algorithm.
The concept-drift detector used is DDM~\citep{DDM}.
All the remaining hyper-parameters were kept at their default values, as defined in River.
\footnote{All the used scripts will also be made public, with instructions to reproduce our experiments after acceptance.}

 
\subsection{Error Measurement}

Table~\ref{tab_all_metrics} shows the RMSE and RMSE$_{\phi}$ results measured while training the different regressors.
The threshold $t_\phi=0.8$ was selected to the relevance of RMSE$_{\phi}$, as previously reported in \cite{aminian2021chebyshev}.
In other words, we are interested in the predictive performance of values whose relevance is higher than $0.8$. We used sliding windows of size $1000$ to compute the metrics.

\begin{table}[!htb]
 \centering
    \small
    \caption{All the measured performance metrics, considering both AMRules and AMRules+ChebyOS. We measure the memory usage and running time relative to the original AMRules. In the case of the number of generated rules, we indicate around parenthesis the percentage of rules whose outputs are higher than $thr_{re}$.}
    \label{tab_all_metrics}
    \begin{tabular}{llrlr}
    \toprule
    \textbf{Metric} & & \multicolumn{1}{c}{\textbf{AMRules}} & & \multicolumn{1}{c}{\textbf{AMRules+ChebyOS}} \\ \midrule
    RMSE                      &  & \textbf{0.9597}    &  & 0.9641     \\
    RMSE$\phi=0.8$            &  & \textbf{3.0869}    &  & 3.0986     \\
    Relative time             &  & \textbf{1.0}       &  & 1.0488     \\
    Relative memory           &  & 1.0       &  & \textbf{0.6783}     \\
    \#Rules ($thr_{re}=2.57$) &  & 4 (0.0\%) &  & \textbf{6 (33.0\%)} \\ \bottomrule
\end{tabular}
\end{table}


Table~\ref{tab_all_metrics} also depicts the impacts of applying the oversampling strategy on the amount of memory consumed by the model and the processing time. We get a negligible difference in the run time and a decrease in the amount of used memory.




As expected, the standard AMRules algorithm yielded the smallest RMSE values compared with the AMRules with the ChebyOS sampling strategy.
RMSE privileges models that produce outputs close to the mean value of the target, which is often the desideratum for regression.
However, in imbalanced regression tasks, the extremes of the target variable distribution are the focus.
In our case, the highest values of reconstruction error are indications of potential failures in the train.


\subsection{ Explanations}
Explainable Artificial Intelligence (XAI) is becoming a game changer in all domains. 
It is specifically designed to address AI systems' transparency and trust issues.
This is especially relevant when signalling critical alarms. At the same time, the system signals an alarm; it must be able to explain why the alarm.
The explanation must be expressed in a way that is understandable to users.

Our systems generate local explanations for specific failures. The intended audience of these explanations is the workers at the workshop involved in repairing the train. The global explanations provide information about the sensors and components more prone to failures. This is information relevant for managers and engineers.

\subsubsection{Global Explanations}

Global Explanations refer to the set of rules learned after processing the dataset. 
The ruleset is expected to evolve due to the incremental nature of AMRules.
We indicate the number of rules each regressor generates in Table~\ref{tab_all_metrics}. We also show 
the percentage of rules whose outputs exceed the reconstruction error threshold ($thr_{re}$).




We present below a subset of rules obtained for the Train dataset when using 
ChebyOS+AMRules.

Rules 0-9 relate to an air leak after the pneumatic control panel. 
Rules 5 and 7 relate to failure of the compressor control system, i.e., the escape valves installed on the APU are opened when the compressor is trying to fill the tanks.


\begin{Verbatim}[frame=leftline,fontsize=\small]
Rule 0: B5_MC > 8.0 and B7_H1 > -2.0 Then 2.30
Rule 1: B3_TP2 <= 5.0 and  MA1_Oil <= 0.9
        and  MA2_Oil > -0.1 and B3_TP2 > 2.2 Then 1.58
Rule 2: B5_MC > 7.9 and  MA1_Oil <= -2.1000 Then 2.25
Rule 3: B3_MC > 2.9 Then 2.67
Rule 4: B4_MC > -0.3 and B7_TP2 <= -0.1 Then 1.9
Rule 5: B2_H1 > 0.7 Then 2.65
Rule 6: B4_MC <= 0.6 and B6_H1 > -2.0 and dig3 > -1.9
        and B3_TP2 <= -1.1 Then 2.73
Rule 7: B1_H1 > 1.5 Then 1.5
Rule 8: B5_MC <= 7.6 and B7_H1 > -1.6 and B6_TP2 <= -0.4 
        Then 1.8
Rule 9: B1_TP2 > -1.3 and B3_TP2 > -1.1 Then 1.80
\end{Verbatim}

Rules 10 and 11 are related to an oil leak on the compressor's motor. 
Larger sensor values {\tt Oil\_Temperature} less oil in the system. The output of the rules indicates the severity of the failure. 
\begin{Verbatim}[frame=leftline,fontsize=\small]
Rule 10: Max_Oil > 0.9 and Med_DV > -3.2 Then 2.64
Rule 11: MA1_Oil <= 1.2 and B4_TP2 <= 0.9 Then 1.75
\end{Verbatim}  

\subsubsection{Local Explanations}

Local explanations refer to the rules that are triggered by an example. When, for a specific example, the reconstruction error of the LSTM-AE exceeds a threshold, the system outputs the rule or rules that were triggered by that example. This section presents illustrative examples of two air leak failures and one explanation of oil leak failure.

\paragraph{Example 1: Air leak failure}


\begin{Verbatim}[frame=leftline,fontsize=\small]
Sample 2001	re=3.83 2/28/2022 15:16
Rule 0: B3_TP2 <= 5.0 and  MA1_Oil <= 0.9
Final prediction: 1.63
\end{Verbatim}

This failure is related to an air leak in the train clients (breaks, suspension, oil, and sand subsystem). A collateral rule fired was the oil temperature due to the compressor engine's constant working load.


\paragraph{Example 2: Air leak failure}


\begin{Verbatim}[frame=leftline,fontsize=\small]
Sample 9167	re=2.82  3/22/2022 11:48
Rule 7: B1_H1 > 1.5 
Final prediction: 1.52
\end{Verbatim}

This failure is related to a failure in the APU's control system caused by a malfunction of a pneumatic control valve. 
The system opens the escape valves (H1) when the compressor tries to fill the tanks.

\paragraph{Example 3: Oil leak failure}


\begin{Verbatim}[frame=leftline,fontsize=\small]
Sample 9049	re=2.744   5/30/2022 12:17
Rule 11: MA1_Oil <= 1.2 and B4_TP2 <= 0.9
Final prediction: 1.61
\end{Verbatim}
This was a severe failure due to an oil leak. The train driver did not receive any alarm to return to maintenance, and the motor was seized.

\subsubsection{Comparison with Shapley Values}

\begin{figure}
\centering
\includegraphics[scale=0.5]{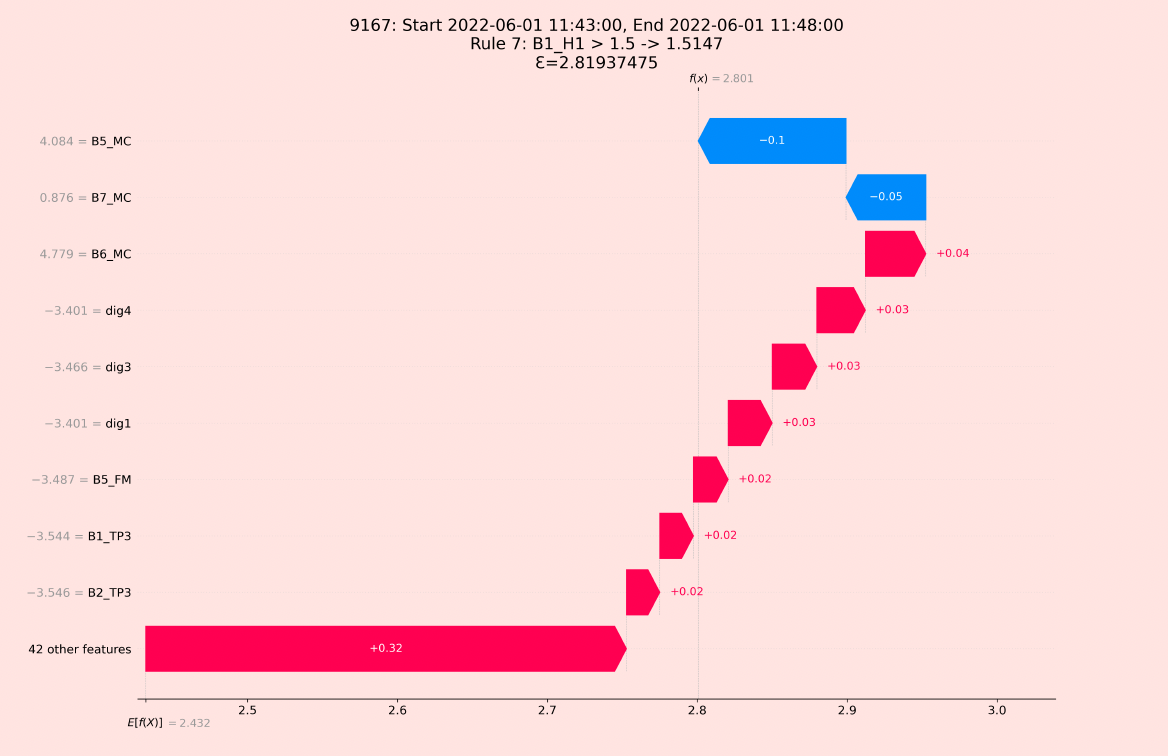}
\caption{Rules versus Shapley values for the air leak failure.}\label{fig:1A}
\end{figure}

Two popular general proposed methods for explainable AI are SHAP values~\cite{nips2017}
(SHapley Additive exPlanations) and LIME~\cite{lime} (Local Interpretable Model-Agnostic Explanations).
They have gained popularity for explaining the decision-making process of machine learning models. 
These frameworks offer instance-level explanations, enabling the analysis of feature contributions for individual predictions.
Figure~\ref{fig:1A} presents the Shapley diagram, referring to the air leak failure. 
The rule that triggered is {\tt If B1\_H1 > 1.5 Then re = 1.52}, with a clear indication of the location of the fault in Valve H1 located at the Compressor module.
While Shapley diagrams always include the (positive or negative) relevance of several features, rules are based on a few literals, which increases interpretability.
Shapley diagrams provide detailed information on each feature's contribution, which might not be advantageous. 
In the 3 failures studied in this paper, the cause of the alarm was correctly identified by the rules.

When providing explanations, we must consider the interlocutor. Local explanations are useful for the maintenance workers at the maintenance workshop to design the repair plan when a failure occurs. Global explanations are useless for interpreting a specific failure but are useful for engineers to understand the critical components of an APU. 

\subsection{Discussion}

The proposed neuro-symbolic architecture explores the best of two worlds: the neural world with high-performance detecting anomalies and the symbolic world that provides interpretable human knowledge.  
The rule model and the neural model are tightly integrated. The rule set is an interpretable model of the behaviour of the LSTM-AE black-box model. When the LSTM-AE signals an alarm for a given input, the rule that triggers that particular input informs about which inputs are relevant for the high value of the reconstruction error. The rule {\bf makes sense of the alarm}: knowing the relevant sensors for the alarm, we know from Table 1 which component has a problem. This is the main contribution of this work: {\bf understanding and giving meaning} to the alarms of black-box models.

In this case study, the rules identified the sensors responsible for the alarm provided by LSTM-AE. The sensor determines the subsystem with problems, which is relevant for the maintenance team at the workshop. 
This way, rules express actionable knowledge and intelligence in a complex environment, directly transforming data into decisions and enabling decision-making actions.

Failures are rare events. With the adoption of Chebyshev sampling, we were able to capture two different failure types. In the first example, we identified the air leak using the H1 and Towers sensors, and in the second example, we captured the oil leak using the Oil Level sensor. None of the failures was captured using AMRules alone.

Finally, to assess the effectiveness of the alarms produced by our model and respective explanations, the engineering and maintenance team carefully analyzed them against the maintenance reports generated by the failures. Our model detected all the failures described in the maintenance reports that led to removing the train from operation. Along with that, the model also reported a significant set of unexpected abnormal APU behaviours that didn't lead to the removal of the train but were indicators of the health condition of the APU. The engineering and maintenance teams agreed that this model allows them to understand specific APU operation behaviours not described by the APU constructor. Moreover, according to them, the model also allowed the conclusion that the equipment often had unexpected and "non-normal" behaviour.


\section{Conclusions}
\label{sec:concl}
Maintenance scheduling and management methodology have become important topics in the industry with the Internet of Things, manufacturing technology, and mass production. In these contexts, there is an urgent need to understand and trust models and their results~\cite{molnar2022}.

One common approach in data-driven predictive maintenance (PdM) is learning the normal operating condition. Any significant deviation from that working condition could be a potential failure. One of the methods to attain this is by learning LSTM-AE from normal examples of operation. The intuition is that if we give an abnormal example of operation, the reconstruction error of the LSTM-AE will be very high. Those examples are relevant from the PdM perspective and are typically rare. This configures an imbalanced data stream regression problem where only very few studies exist.  

In this paper, we propose an online system able to learn in parallel an LSTM-AE trained to identify anomalies and a regression rules algorithm that models the large values of the reconstruction error of LSTM-AE.
The proposed neuro-symbolic model generates 
local and global explanations that play different roles from the maintenance point of view. Local explanations should be used for the repair plan in the presence of a failure signal. Local explanations {\it makes sense} of a failure signal. Global explanations provide information for asset management by identifying critical points in the equipment. 
The main contribution is that the rule learner algorithm is wrapped in a sampling schema, allowing for selecting relevant examples. That is the main reason why the proposed framework works. 
We applied this methodology in a Predictive Maintenance scenario with data from Metro do Porto.
Results showed that we could learn rules for the high values of the reconstruction error. These are the cases of interest. 

The methodology we present is general enough to be applied to other online imbalanced streaming scenarios that use black-box models to predict peaks or bursts in events. We plan to explore different scenarios outside PdM applications for this methodology.
Finally, interpretability and explainability are human-computer-interface problems. While decision rules have a high degree of understandability for computer-science people, we will need to translate the rules into a natural language to enlarge the scope of people that understand the messages. That is left for future work.

\section*{Acknowledgments}
This work was supported by the CHIST-ERA grant CHIST-ERA-19-XAI-012 and project CHIST-ERA/0004/2019, funded by FCT.
Saulo Mastelini also acknowledges the support of FAPESP grant 2021/10488-7.




\end{document}